\documentclass[10pt,twocolumn,letterpaper]{article}

\usepackage{cvpr}              

\definecolor{cvprblue}{rgb}{0.21,0.49,0.74}
\usepackage[pagebackref,breaklinks,colorlinks,allcolors=cvprblue]{hyperref}
\usepackage{bm} 
\usepackage{graphicx}
\usepackage{multirow} 
\definecolor{darkgreen}{rgb}{0.0, 0.5, 0.0} 

\newcommand{\pparagraph}[1]{\medskip\noindent\textbf{#1}}

\def\paperID{3129} 
\def\confName{CVPR}
\def\confYear{2025}

\title{Structure Tensor Representation for Robust Oriented Object Detection}

\author{Xavier Bou\textsuperscript{1} \and Gabriele Facciolo\textsuperscript{1} \and Rafael Grompone von Gioi\textsuperscript{1} \and Jean-Michel Morel\textsuperscript{2} \and Thibaud Ehret\textsuperscript{1}\\
\textsuperscript{1}Université Paris-Saclay, CNRS, ENS Paris-Saclay, Centre Borelli, France \\
\textsuperscript{2}City University of Hong Kong, Department of Mathematics, Kowloon, Hong Kong \\
{\tt\small xavier.bou\_hernandez@ens-paris-saclay.fr}
}

\begin{document}
\maketitle
\begin{abstract}
Oriented object detection predicts orientation in addition to object location and bounding box. Precisely predicting orientation remains challenging due to angular periodicity, which introduces boundary discontinuity issues and symmetry ambiguities. Inspired by classical works on edge and corner detection, this paper proposes to represent orientation in oriented bounding boxes as a structure tensor. This representation combines the strengths of Gaussian-based methods and angle-coder solutions, providing a simple yet efficient approach that is robust to angular periodicity issues without additional hyperparameters. Extensive evaluations across five datasets demonstrate that the proposed structure tensor representation outperforms previous methods in both fully-supervised and weakly supervised tasks, achieving high precision in angular prediction with minimal computational overhead. Thus, this work establishes structure tensors as a robust and modular alternative for encoding orientation in oriented object detection. We make our code publicly available, allowing for seamless integration into existing object detectors.
\end{abstract}

\section{Introduction}
\label{sec:intro}
\begin{figure}[t]
    \centering
    \includegraphics[width=1\linewidth,trim={0cm 7.2cm 15.1cm 0.8cm},clip]{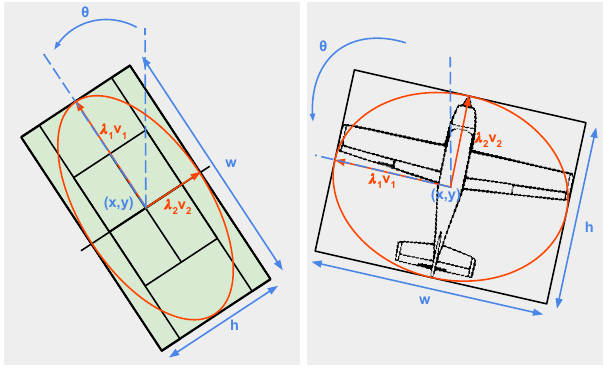}
    \caption{\textbf{The proposed angle representation}. Comparison between the traditional oriented bounding box format $(x, y, w, h, \theta)$ (in blue) and our structure tensor representation $T$ (in orange). Orientation and anisotropy are represented in T by its eigenvalues $\lambda_1$ and $\lambda_2$, and their corresponding eigenvectors $v_1, v_2$.\vspace{-6mm}}
    \label{fig:teaser}
\end{figure}

Object detection is one of the classical problems in computer vision and traditionally localizes objects with horizontal bounding boxes (HBB)~\cite{obj_det_review_1}. Nevertheless, certain areas such as aerial imagery~\cite{dota, fair1m, oriented_detection_1, h2rbox, h2rbox-v2} and scene text detection~\cite{scene_text_det_1, scene_text_det_2, scene_text_det_3, scene_text_det_4} require information regarding the orientation of objects for more accurate identification~\cite{mmrotate}. Hence, oriented object detection extends object detection by predicting  oriented bounding boxes (OBB) that better align with object boundaries.

Despite progress in recent years, angular periodicity still poses challenges when predicting object orientation. In the first place, small variations between prediction and ground truth at the angular boundary will cause a sharp loss increase. This phenomenon, known as the boundary problem~\cite{theoretically_achieving, csl, xu2024rethinking}, can penalize the network while comparing mathematically similar values. In addition, objects with axial symmetry can cause the network to penalize equivalent angle predictions. For example, $\frac{\pi}{2}$ and $-\frac{\pi}{2}$ radians are equivalent orientations for rectangular objects, but their loss would be significantly high~\cite{theoretically_achieving, psc}.

Previous works have addressed these discontinuity problems in different manners. Some approaches propose angle-coder solutions~\cite{csl, dcl, psc}, which transform orientation into representations that circumvent angular periodicity challenges. These are robust and modular, yet they may require extensive hyperparameter tuning. Other methods mitigate these issues by representing OBBs as 2D Gaussian distributions~\cite{wasserstein_loss, kld, kfiou}, and compare the distance between these representations through distribution-based computations. They elegantly provide a continuous orientation representation though introduce complexity and may lack stability. Furthermore, reported improvements focus on mean average precision (mAP), without specifically evaluating the precision of orientation predictions.

Traditional image analysis methods, particularly for edge and corner detection~\cite{harris1988combined, forstner1987fast, lindeberg1998edge, brox2006level, scharr2004optimal}, have investigated the representation of orientation, shape, and symmetry via the concept of structure tensor~\cite{structure_tensor}. A structure tensor is a 2x2 symmetric matrix that encodes information of the orientation and anisotropy of a local structure. They offer two key advantages: 1) the representation of orientation without boundary discontinuities and 2) the flexibility to model different symmetries through their eigenvalues. Moreover, calculating structure tensors is straightforward and computationally efficient, which can be done in an angle-coder fashion. Thus, as illustrated in Figure~\ref{fig:teaser}, we propose to represent object orientations as structure tensors, combining the strengths of angle-coder methods and Gaussian-based approaches.

We perform extensive experiments and compare the performance of our approach to previous works. We show that the structure tensor representation outperforms state-of-the-art (SOTA) methods in many cases across five different datasets on mAP. Furthermore, complementary ablations justify choices made and show that our approach yields strong precision in angular prediction at a low computational cost. Our contributions can be summarized as:
\begin{itemize}
  \item To the best of out knowledge, this is the first work to propose to represent orientation in OBBs as structure tensors. A simple and modular implementation is made publicly available, enabling its integration into any detector.
  \item Through extensive experimentation, we show that our approach consistently achieves SOTA performance and even outperforms existing methods in many cases.
  \item We combine the strengths of angle-coder and Gaussian-based methods, resulting in a robust, computationally efficient solution that elegantly addresses boundary problems without requiring additional parameters.
\end{itemize}
\section{Related works}
\label{sec:formatting}
\paragraph{Object Detection} is one of classical problems in computer vision that aims to identify instances of visual objects in images and precisely estimate their location, providing an important piece of information: \textit{What objects are where?}~\cite{OD_review, obj_det_review_1}. Similarly to other fields, the advances in deep neural networks (DNN) represented large improvements in object detection, and DNN-based approaches became the state of the art over classical approaches~\cite{obj_det_review_2}. Modern object detectors can be grouped into one-stage detectors and two-stage detectors. One-stage detectors localize potential objects over a grid of locations, and determine their category in one single step~\cite{yolo, retinanet, mobilenet}. Instead, two-stage detectors first extract a number of class-agnostic bounding box proposals using a region proposal network. Then, a sub-network classifies these region proposals amongst all possible object categories~\cite{rcnn, fasterrcnn, maskrcnn}. While one-stage detectors provide a more time-efficient solution by merging the regression and classification problems into one single step, two-stage detectors generally achieve higher performances. Regardless of their type, early DNN-based detectors considered a dense sampling of locations, known as anchors, to predict bounding boxes in the image. However, later works proposed anchor-free approaches in order to achieve the performance of two-stage detectors without the computational cost of anchor-based regression~\cite{cornernet, centernet, fcos, reppoints}. More recently, Carion \textit{et al.} introduced DETR~\cite{detr}, a transformer-based method that detects objects using a fixed set of learnable queries. The concept of query-based detection has had considerable impact in the field, inspiring a progression of recent works~\cite{detrs_1, detrs_2,detrs_3,detrs_4,detrs_5,detrs_6}.

\paragraph{Oriented Object Detection (OOD).} Traditional object detection localizes objects via horizontal bounding boxes, i.e. boxes aligned with the $(x, y)$ axes. However, in some particular domains such as aerial imagery~\cite{dota, fair1m, oriented_detection_1, h2rbox, h2rbox-v2} or scene text detection~\cite{scene_text_det_1, scene_text_det_2, scene_text_det_3, scene_text_det_4}, extracting the orientation of objects is preferred~\cite{mmrotate}. Thus, OOD predicts an oriented bounding box $(x, y, w, h, \theta)$ instead of the horizontal bounding box $(x, y, w, h)$, where $(x, y)$ corresponds to the center position, $(w, h)$ is the width and height, and $\theta$ indicates the rotation angle~\cite{xu2024rethinking}. To achieve this, several works adapted popular object detection architectures to handle the additional angle prediction~\cite{oriented-rcnn, rot_transf, R3det, Redet, rot_fcos, gliding_vertex, yi2021oriented}. More recently, Yang \textit{et al.}~\cite{h2rbox} developed an approach that learns to detect oriented objects via horizontal bounding box supervision, thus considerably decreasing the annotation cost. Their method, coined H2RBox, leverages rotational covariance to find the minimum circumscribed rectangle inside the HBB. An improved version H2RBox-v2~\cite{h2rbox-v2} was later introduced, addressing angular periodicity and achieving OBB-supervised SOTA performance. Moreover, some works have proposed to learn oriented detection via single point supervision, where only a point is provided for each object. However, the performance of these methods is still far behind HBB and OBB-supervised approaches~\cite{point2rbox, pointobb}.

\paragraph{Boundary and Symmetry Problems.} The angle regression task OOD introduces two well-known challenges:

1) The \textbf{boundary problem}~\cite{theoretically_achieving, csl, xu2024rethinking} refers to the rotation discontinuities that occur due to angular periodicity. Small differences between prediction and ground truth at the angular boundary will cause a sharp loss increase during training, despite both values being mathematically similar. 

2) \textbf{Symmetry issues}~\cite{theoretically_achieving, psc}, caused by rectangular and squared-like objects, can also negatively impact the learning process. In the case of rectangular objects, a rotation of $\frac{\pi}{2}$ or $-\frac{\pi}{2}$ radians yields equivalent bounding boxes. Nevertheless, the training will penalize a network that correctly predicts $-\frac{\pi}{2}$ if the ground truth is $\frac{\pi}{2}$. Squared objects behave similarly with a $\frac{\pi}{2}$ periodicity instead of $\pi$. This incoherence between loss and box alignment is a source of confusion for the network during training.

Mainly three strategies have been proposed to mitigate these problems:

a) \textbf{Smoothing losses}: These works aim to smooth the loss discontinuities at the angular boundary. SCRDet~\cite{scrdet} uses smooth L1-loss, while RSDet~\cite{rsdet} uses a modulated loss. SCRDet++~\cite{scrdet++} introduced an IoU factor into the loss term to further address boundary issues.

b) \textbf{Angle-coder} methods transform object orientations into representations that are robust to angular periodicity issues. Yang and Yan~\cite{csl} re-framed angle prediction as a classification task, representing orientation using a Circular Smooth Label (CSL) to address the error between adjacent angles. Densely Coded Labels (DCL)~\cite{dcl} later improved performance and speed with a new re-weighting loss based on aspect ratio and angle distance, and addressed symmetric ambiguities of square objects. Yu and Da~\cite{psc} transformed rotational periodicity of different cycles into the phase of different frequencies. Their approach, Phase-Shifting Coder (PSC), proved to be robust and surpassed previous angle representations as a regression task. While angle-coder methods are modular and straightforward, they often require extensive hyperparameter tuning.

c) \textbf{Gaussian-Based Methods}: These approaches model OBBs as 2D Gaussian distributions to solve boundary and symmetry challenges. The Gaussian Wasserstein Distance (GWD)~\cite{wasserstein_loss} compares predicted and ground-truth distributions, offering an approximation to the non-differentiable IoU loss. Yang \textit{et al.}~\cite{kld} further refined this with the Kullback-Leibler Divergence (KLD) and later introduced KFIoU~\cite{kfiou}, a fully differentiable method that approximates SkewIoU and handles non-overlapping cases without parameters. Gaussian-based approaches provide an elegant solution to boundary discontinuities and symmetric ambiguities, though they may lack stability and introduce complexity due to their distribution-based losses.

\section{Motivation}
Previous works have attempted to circumvent angular discontinuity issues in different manners. Gaussian-based techniques offer a continuous representation of orientation, but their performance may be inconsistent across datasets, and they introduce the complexity of distribution-based computations. In contrast, angle-coder methods provide a modular and robust solution, though still struggle with continuity and may require extensive hyperparameter tuning to manage these challenges.

\paragraph{The Structure Tensor} is a classic concept in image processing and computer vision introduced by Bigun \textit{et al.}~\cite{structure_tensor}, extensively used in corner and edge detection~\cite{harris1988combined, forstner1987fast, lindeberg1998edge, brox2006level, scharr2004optimal}. Also known as the second-moment matrix, a structure tensor is a low-level feature represented as a 2D symmetric matrix that describes orientation and anisotropy in a local neighborhood. The structure matrix $J$ at each pixel is defined as
\begin{equation}\label{eq:structure_tensor_def}
     J = \begin{bmatrix}
                    J_{x}^2 & J_{x}J_y \\
                    J_{x}J_y & J_{y}^2
    \end{bmatrix},
\end{equation}
where $J_{x}$ and $J_{y}$ are the gradients of the image in the x and y directions, respectively. Let $\lambda_1, \lambda_2$ be the eigenvalues of $J$ with $\lambda_1 \geq \lambda_2$, and $v_1, v_2$ 
corresponding eigenvectors. The eigenvectors of $J$ represent the principal directions of the local gradients, where $v_1$ points in the direction of the main orientation of the local intensity pattern, and $v_2$ is orthogonal to it. Moreover, the eigenvalues represent the level of intensity variation in the principal directions, and their relative difference encapsulates the anisotropy of the structure, such that: 

\begin{itemize}
  \item If $\lambda_1 \gg \lambda_2$, the local structure is highly anisotropic, thus the gradients are much stronger in one direction.
  \item If $\lambda_1 \approx \lambda_2$, the local structure is isotropic, i.e. the gradients are similar in all directions.
  \item If $\lambda_1 \approx 0 $ and $ \lambda_2 \approx 0$ , the structure is homogeneous.
\end{itemize}

Structure tensors combine the advantages of both Gaussian-based and angle-coder approaches. Like Gaussian-based methods, they offer a continuous orientation representation that adapts to different symmetries through the relation between their eigenvalues. Moreover, structure tensors enable a computationally efficient representation without the need for hyperparameters, which can be implemented in an angle-coder fashion. Thus, representing OBBs as structure tensors not only addresses the challenges of angular periodicity and symmetry but additionally has the potential to simplify the orientation prediction task.
\section{Method}
\label{sec:method}
\begin{figure*}[t]
    \centering
    \includegraphics[width=1\linewidth,trim={0.1cm 8.05cm 
    9cm 0.2cm},clip]{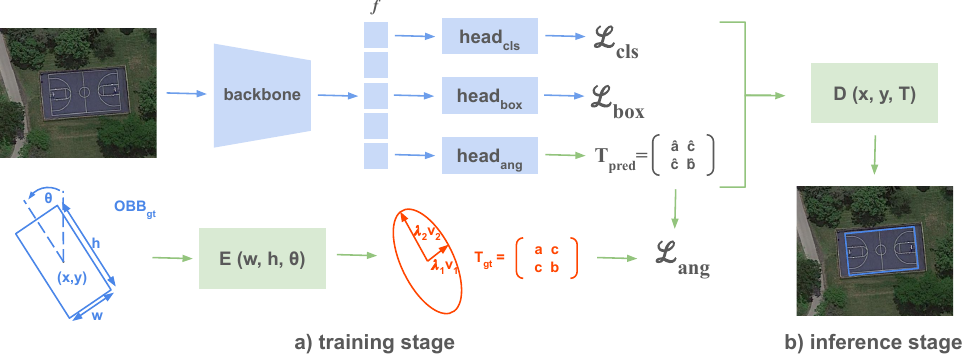}
     \caption{\textbf{Structure tensor representation in a neural network}. (a) During training, the backbone extracts image features $f$, which are used for classification and regression. The angle head predicts orientation as a structure tensor $T_{pred}$, and the ground truth $OBB_{gt}$ is encoded into $T_{gt}$ for angle loss computation. (b) At inference, $T_{pred}$ is decoded into the standard OBB format $(x,y,w,h,\theta)$. Blue denotes standard detector components, while green highlights our method.\vspace{-6mm}}
    \label{fig:general_diagram}
\end{figure*}
The key idea of this paper is to address angular boundary discontinuities and symmetry issues in OOD by representing the orientation of objects through structure tensors.

Let $o_{bb}$ be an oriented bounding box characterized by the usual 5-parameter format $(x, y, w, h, \theta)$, where $(x,y)$ are the center coordinates, $(w, h)$ are the width and height, and $\theta$ is the rotation angle.  We introduce the encoding and decoding functions, $E$ and $D$, which map the orientation of the OBB into a structure tensor T and recover the OBB from T, respectively:

\begin{equation}\label{eq:encode_func}
o_{bb} \xrightarrow{E(w, h, \theta)} T \xrightarrow{D(x, y, T)} o_{bb}.
\end{equation}
Thus, we propose to directly predict orientation in the structure tensor space to circumvent angular periodicity issues in an angle-coder manner. Figure~\ref{fig:general_diagram} provides a comprehensive overview of our approach integrated into a oriented object detector. The upcoming text describes in detail the encoding and decoding transformations, as well as how to integrate them in current-day detectors.

\paragraph{Encoding.} The function $E$ maps an OBB to a structure tensor $T$ that encapsulates the object orientation in a continuous manner and reflects the anisotropy of the object, granting the flexibility to handle axial symmetry. To do so, the eigenvalues $\lambda_1$ and $\lambda_2$ of $T$ represent the height and width of the OBB, respectively, and their eigenvectors $v_1=[v_{11}, v_{12}]$ and $v_2=[v_{21}, v_{22}]$ depict the orientation of the object $\theta$. To this end, we first define the rotation matrix $R_{\theta}$ that denotes the rotation by $\theta$ radians, and the diagonal matrix $\Lambda$ that depends on the width and height of the bounding box:
\begin{equation}\label{eq:rotation_diag_matrices}
\begin{array}{l}
R_{\theta} = \begin{bmatrix}
                    \cos{\theta} & -\sin{\theta} \\
                    \sin{\theta} & \cos{\theta}
    \end{bmatrix},\; \Lambda = \begin{bmatrix}
\frac{w}{2} & 0 \\
0 & \frac{h}{2}
\end{bmatrix}.
\end{array}
\end{equation}
The structure tensor $T$ is then computed as
\begin{equation}\label{eq:structure_tensor}
T = R_{\theta} \Lambda R_{\theta}^T = \begin{bmatrix}
                    a & c \\
                    c & b
    \end{bmatrix},
\end{equation}
where $T$ is parameterized by three values $a$, $b$, and $c$. We draw similarities to Gaussian-based approaches, e.g. GWD~\cite{wasserstein_loss}, KLD~\cite{kld} or KFIoU~\cite{kfiou}. These methods represent OBBs as 2D Gaussian distributions where the covariance matrix $\Sigma^{1/2}$ is computed comparably to our structure tensor. Nevertheless, we do not utilize Gaussian distributions nor compare them via Kullback–Leibler divergence or Wasserstein distance. We simply apply a geometric transformation and directly regress the three parameters ($a$, $b$, $c$) of the structure tensor, allowing us to encode and decode the orientation of objects in a more straightforward manner without the need for complex distribution-based metrics.

\paragraph{Decoding.} Given a structure tensor $T$ and the center coordinates $(x, y)$, we want to decode $T$ back to the original $o_{bb}$. First, we compute the eigenvalues $\lambda_1$ and $\lambda_2$ and eigenvectors $v_1$ and $v_2$ of $T$. As mentioned earlier, the eigenvalues of $T$ characterize the width and height of the bounding box. Furthermore, the orientation of the object is depicted by the direction of the strongest eigenvalue. All in all, we can extract the $w$, $h$ and $\theta$ as
\begin{equation}\label{eq:decode_wh}
\begin{array}{l}
w = {2\lambda_1},\:\: h = {2\lambda_2},\:\: \theta = \arctan(\frac{v_{12}}{v_{11}}),
\end{array}
\end{equation}
where $v_{11}$ and $v_{12}$ are elements of the eigenvector $v_1$ associated with the larger eigenvalue $\lambda_1$. With known (x,y) values, we can just provide the decoded $o_{bb}$ as $(x, y, w, h, \theta)$.

\pparagraph{Training.} Similarly to Yu and Da~\cite{psc}, we propose an angle-coder approach where the regression of the angle $\theta$ is learned by the network in the structure tensor space. In the traditional two-stage detector architecture, a backbone $\mathcal{F}$ extracts high-dimensional features $f$ from an image $I \in \mathbb{R}^{3 \times H \times W}$:
\begin{equation}\label{eq:feature_extraction}
     \mathcal{F}(I) = f \in \mathbb{R}^{H' \times W' \times D},
\end{equation}
where $H$ and $W$ are the image height and width, $H'$ and $W'$ are the height and width of $f$ with the common decrease in spatial resolution in DNNs, and $D$ is the feature dimensionality. Then, different heads are applied to $f$ for the regression and classification of objects, where typically $head_{cls}$ is tasked with classification, $head_{bbox}$ regresses a horizontal bounding box, and $head_{ang}$ predicts the orientation. 

In this context, we integrate the proposed representation into network training by letting $head_{ang}$ predict the three parameters $\hat{a}, \hat{b}, \hat{c}$ that characterize the predicted structure tensor $\mathbf{T}_{pred}$. During training, the ground truth oriented bounding boxes are encoded into structure tensor representations via the encoder $E$. Let $\mathbf{OBB}_{gt}$ represent the ground truth bounding box and $\mathbf{T}_{gt}$ its corresponding structure tensor, the angle loss $\mathcal{L}_{ang}$ is then computed by the L1-Loss between the predicted and ground truth structure tensors:

\begin{equation}\label{eq:l1_loss}
     \mathcal{L}_{ang} = \frac{1}{N} \sum_{i=1}^{N} \left| \mathbf{T}_{gt}^i - \mathbf{T}_{pred}^i \right|
\end{equation}
where $N$ is the total number of bounding boxes in the training batch. Then, the overall loss is computed as the traditional three-term loss:
\begin{equation}\label{eq:total_loss}
     \mathcal{L} = w_{cls} \mathcal{L}_{cls} + w_{bbox} \mathcal{L}_{bbox} + w_{ang} \mathcal{L}_{ang},
\end{equation}
where $w_{cls}$, $w_{bbox}$, and $w_{ang}$ are the weights for the classification, bounding box regression, and angle regression losses, respectively.
\begin{figure}
    \centering
    \includegraphics[width=1\linewidth,trim={0cm 0cm 0cm 0cm},clip]{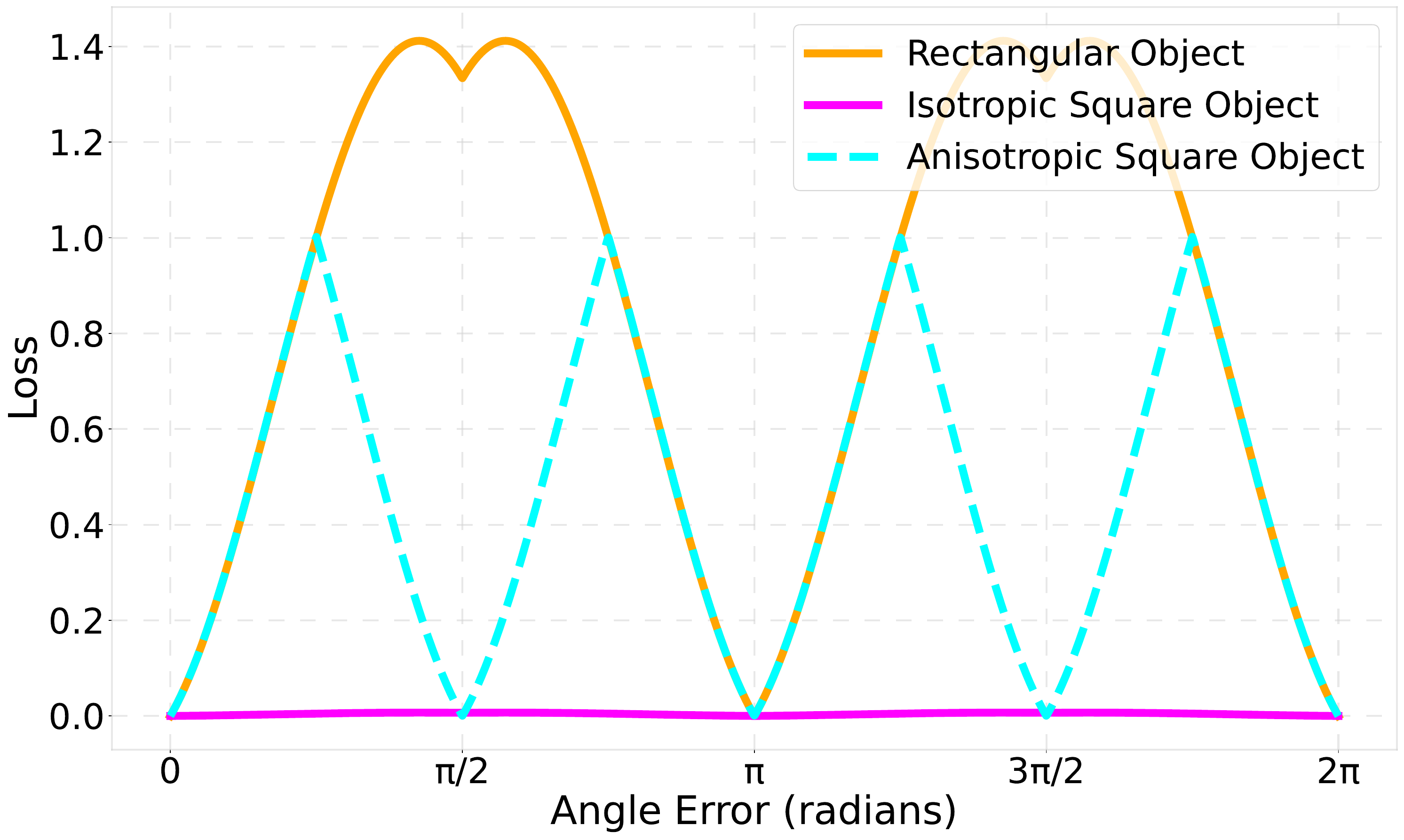}
   \caption{\textbf{Loss behavior with isotropic structure tensors}. Rectangular objects exhibit $\pi$ periodicity, with loss increasing as angle error grows, while isotropic (square-like) objects yield consistently low loss regardless of angle error. Introducing anisotropy results in a loss with $\frac{\pi}{2}$ periodicity that yields higher values as prediction error increases.\vspace{-6mm}}
    \label{fig:loss_plot_isotropic_issues}
\end{figure}

\pparagraph{Handling Isotropic Objects.} Isotropic structures present the same properties in all directions and therefore lack a unique principal direction. In our structure tensor representation, this is characterized by equal eigenvalues $\lambda_1 = \lambda_2$, corresponding to a square-shaped OBB where $w = h$. The absence of principal orientation is problematic for oriented object detection because it prevents to extract a unique orientation for squared objects. Assuming the OBB format $(x, y, w, h, \theta)$ represents the width as the longest side, i.e. $w \geq h$, a straightforward way to circumvent this issue would be to add a small, negligible value to $w$ when $w \approx h$. This breaks the isotropy of the object without introducing significant error, ensuring the representation of a principal direction. However, when $w \approx h$, the loss between prediction and ground truth structure tensors will be small regardless of the encoded angle, thus leading the network to predict a random orientation for squared-like objects. This can be observed in Figure~\ref{fig:loss_plot_isotropic_issues}, where the L1-Loss between isotropic tensors representing square objects is considerably low for any angle error.

Instead, we handle the cases where $w = h$ by adjusting the structure tensor representation so that $\lambda_1 = w$ and $\lambda_2 = \frac{h}{2}$. To maintain the $\frac{\pi}{2}$ symmetry of square-like objects, the encoder normalizes the angle within the $\frac{\pi}{2}$ range when $|w - h| \leq \epsilon$, where $\epsilon$ is a small threshold that determines when an OBB can be considered square. This approach avoids the absence of a principal direction in the structure tensor representation and ensures a higher loss when the prediction angle deviates from the ground truth. A visualization of the loss behavior is provided in Figure~\ref{fig:loss_plot_isotropic_issues}, and an ablation with different levels of anisotropy is provided later in Section~\ref{sec:experiments}.


\section{Experiments}
\label{sec:experiments}
\begin{figure*}[t]
    \centering
    \includegraphics[width=\linewidth,trim={0.5cm 4.55cm 7.6cm 0cm},clip]{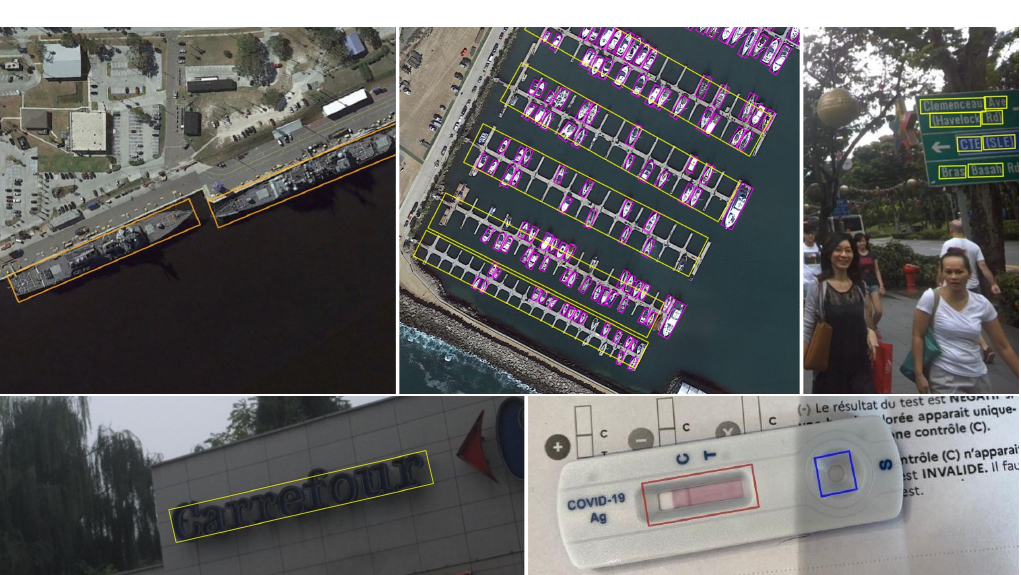}
    \caption{\textbf{Qualitative results of the proposed approach on several datasets.} On the top, from left to right, detection examples from HRSC2016, DOTA, and ICDAR2015 are shown. On the bottom, the left image corresponds to the MSRA-TD500 dataset, while the one on the right belongs to the C19TD test dataset.}
    \label{fig:qualitative_results}
\end{figure*}
\begin{table*}[t]
\centering
\begin{tabular}{ccccccc}  \cmidrule[0.4pt](lr{0.125em}){2-7}%

 \textbf{} & \textbf{Method} & \textbf{Architecture} & \multicolumn{2}{c}{\textbf{DOTAv1.0}} & \multicolumn{2}{c}{\textbf{HRSC2016}} \\ \cmidrule[0.4pt](lr{0.125em}){4-5} \cmidrule[0.4pt](lr{0.125em}){6-7}

\textbf{} & \textbf{} &  & \textbf{mAP50} & \textbf{mAP50:95} & \textbf{mAP50} & \textbf{mAP50:95} \\
\cmidrule[0.4pt](lr{0.125em}){4-4}%
\cmidrule[0.4pt](lr{0.125em}){5-5}%
\cmidrule[0.4pt](lr{0.125em}){6-6}%
\cmidrule[0.4pt](lr{0.125em}){7-7}%
\multirow{8}{*}{\rotatebox[origin=c]{90}{\shortstack{\textbf{OBB}\\\textbf{supervised}}}} 
& Rotated RetinaNet~\cite{retinanet} & RetinaNet & 69.6 & \textbf{\textcolor{darkgreen}{40.9}} & 83.7 & 52.8 \\
& Rotated FCOS~\cite{fcos} & FCOS & 71.1 & \textbf{\textcolor{blue}{40.6}} & 89.0 & 64.0 \\
& CSL~\cite{csl} & FCOS  & \textbf{\textcolor{blue}{71.4}} & 40.5 & 89.5 & 58.4 \\
& GWD~\cite{wasserstein_loss} & RetinaNet & 60.3 & 33.9 & 85.7 & 54.4 \\
& KLD~\cite{kld} & FCOS & \textbf{\textcolor{darkgreen}{71.6}} & 39.6 & 89.8 & 63.6 \\
& KFIoU~\cite{kfiou} & RetinaNet & 69.5 & 37.6 & 84.6 & 48.5 \\
& PSC~\cite{psc} & FCOS & 70.2 & 39.2 & \textbf{\textcolor{blue}{90.0}} & \textbf{\textcolor{blue}{67.6}} \\
& \textbf{ST (Ours)} & FCOS & 71.3 & 39.5 & \textbf{\textcolor{darkgreen}{90.3}} & \textbf{\textcolor{darkgreen}{67.7}} \\  \cmidrule[0.4pt](lr{0.125em}){2-7}%
\multirow{3}{*}{\rotatebox[origin=c]{90}{\shortstack{\textbf{HBB}\\\textbf{supervised}}}}
& H2RBox-v2 w/ CSL & FCOS  & 43.9 & 15.5 & 0.80 & 0.37 \\
& H2RBox-v2 w/ PSC\cite{h2rbox-v2} & FCOS  & \textbf{\textcolor{blue}{71.5}} & \textbf{\textcolor{darkgreen}{39.4}} & \textbf{\textcolor{blue}{89.1}} & \textbf{\textcolor{blue}{57.1}} \\
& \textbf{H2RBox-v2 w/ ST (Ours)} & FCOS  & \textbf{\textcolor{darkgreen}{72.6}} & \textbf{\textcolor{blue}{39.2}} & \textbf{\textcolor{darkgreen}{89.5}} & \textbf{\textcolor{darkgreen}{59.0}} \\  \cmidrule[0.4pt](lr{0.125em}){2-7}%
\end{tabular}
\caption{\textbf{Results on the remote sensing datasets DOTAv1.0 and HRSC}, showing mAP50 and mAP50:95 for the proposed structure tensor representation and compared to SOTA methods. Both OBB-supervised and HBB-supervised approaches are reported. For each metric, the best score and the second best score are shown in green and blue, respectively.\vspace{-6mm}}
\label{table:dota_hrsc_results}
\end{table*}
\begin{table*}[t]
\centering
\begin{tabular}{cccccc}  \cmidrule[0.4pt](lr{0.125em}){2-6}%

 \textbf{} & \textbf{Method} & \textbf{Architecture} & \textbf{ICDAR15} & \textbf{MSRA-TD500} & \textbf{C19TD} \\ \cmidrule[0.4pt](lr{0.125em}){4-4} \cmidrule[0.4pt](lr{0.125em}){5-5} \cmidrule[0.4pt](lr{0.125em}){6-6}
\cmidrule[0.4pt](lr{0.125em}){4-4}%
\cmidrule[0.4pt](lr{0.125em}){5-5}%
\cmidrule[0.4pt](lr{0.125em}){6-6}%
\multirow{8}{*}{\rotatebox[origin=c]{90}{\shortstack{\textbf{OBB}\\\textbf{supervised}}}} 
& Rotated RetinaNet~\cite{retinanet} & RetinaNet & \textbf{\textcolor{blue}{65.7}}	& 64.6 & 95.2\\
& Rotated FCOS~\cite{fcos} & FCOS & 60.4	& 67.9 & 94.9\\
& CSL~\cite{csl} & FCOS  & 55.7	& 68.2 & 90.5\\
& GWD~\cite{wasserstein_loss} & RetinaNet & 65.5 & 64.8 & 93.7\\
& KLD~\cite{kld} & FCOS & 64.4	& 68.5 & \textbf{\textcolor{blue}{95.4}}\\
& KFIoU~\cite{kfiou} & RetinaNet & 65.3	& 64.4 & 95.3\\
& PSC~\cite{psc} & FCOS & 64.2	& \textbf{\textcolor{blue}{68.7}}& 94.2\\
& \textbf{ST (Ours)} & FCOS &  \textbf{\textcolor{darkgreen}{66.8}}	& \textbf{\textcolor{darkgreen}{69.0}}  & \textbf{\textcolor{darkgreen}{95.5}}\\
\cmidrule[0.4pt](lr{0.125em}){2-6}%

\multirow{3}{*}{\rotatebox[origin=c]{90}{\shortstack{\textbf{HBB}\\\textbf{supervised}}}}
& H2RBox-v2 w/ CSL & FCOS & \textbf{\textcolor{blue}{41.4}} & 9.1 & 81.4 \\
& H2RBox-v2 w/ PSC\cite{h2rbox-v2} & FCOS & 41.0 & \textbf{\textcolor{blue}{24.0}} & \textbf{\textcolor{darkgreen}{95.2}} \\
& \textbf{H2RBox-v2 w/ ST (Ours)} & FCOS & \textbf{\textcolor{darkgreen}{46.2}} & \textbf{\textcolor{darkgreen}{32.0}} & \textbf{\textcolor{blue}{95.0}} \\
\cmidrule[0.4pt](lr{0.125em}){2-6}%
\end{tabular}
\caption{\textbf{Results on the scene text detection benchmarks ICDAR15 and MSRA-TD500}, showing mAP50 for the proposed structure tensor representation and compared to SOTA methods. Both OBB-supervised and HBB-supervised approaches are reported. For each metric, the best score and the second best score are shown in green and blue, respectively.\vspace{-6mm}}
\label{table:text_and_covid_results}
\end{table*}

\subsection{Benchmarks and implementation details}
We evaluate the proposed approach on different domains. Firstly, we select two extensively used satellite imagery datasets, namely DOTA~\cite{dota} and HRSC2016~\cite{hrsc2016}. Then, we evaluate on ICDAR2015~\cite{icdar2015} and MSRA-TD500~\cite{msra}, two scene text detection benchmarks. Moreover, we introduce a novel dataset containing SARS-CoV-2 (COVID-19) tests.

\pparagraph{DOTA}~\cite{dota}: DOTAv1.0 is a popular remote sensing benchmark with 2,806 images that contain 188,282 object annotations. Instances are classified into 15 categories, some of which include \textit{plane}, \textit{tennis court} and \textit{small vehicle}.

\pparagraph{HRSC2016}~\cite{hrsc2016}: The HRSC2016 dataset contains instances of ships in different orientations, both at sea and near land. The training, validation and test sets include 436, 181 and 444 images of different sizes.

\pparagraph{ICDAR2015}~\cite{icdar2015}: The ICDAR2015 is an oriented scene text detection benchmark that contains 1,000 training images and 500 testing images.

\pparagraph{MSRA-TD500}~\cite{msra}: The MSRA-TD500 is an oriented scene text detection dataset with 300 training images and 200 testing images. It contains English and Chinese text annotated at the sentence level.

\pparagraph{C19TD}: We introduce the COVID-19 Test Dataset (C19TD), which contains cellphone-made images of tests to diagnose SARS-CoV-2 (COVID-19). The goal of the dataset is to localize two test landmarks that are essential to determine the outcome and validity of the test: the \textit{well}, i.e. the region where the reactive chemical is applied, and the \textit{result} area, which indicates the outcome of the test. We highlight the significance of this data as it contains noticeably symmetric objects that do not belong to the remote sensing domain. The training, validation and test splits contain 800, 100 and 102 images of COVID tests, respectively. For robustness, we randomly add unrelated natural images without annotations, expanding the train, val and test sets to 1199, 150 and 150, respectively.

\paragraph{Experimental setup.} We evaluate and compare the structure tensor representation against other SOTA methods. To this aim, we implement our method as an angle-coder in MMRotate~\cite{mmrotate}, and train a FCOS object detector with a ResNet-50 backbone using structure tensors. To make comparisons fair, we train SOTA methods on the same architecture and, when not possible, we keep the ResNet-50 backbone. In addition, we train and compare HBB-supervised methods using FCOS with ResNet-50 using different angle representations. HBB-supervised approaches are weakly supervised with horizontal bounding boxes and lack orientation ground truth. All models are trained on a NVIDIA RTX 6000 with a batch size of 2. For the DOTA dataset, we apply the standard pre-processing as per MMRotate~\cite{mmrotate}, generating image crops of 1024x1024 with an overlap of 200 pixels. The remaining datasets are pre-processed by resizing images to 800x512 resolution and augmented with random flips and random rotations. We respect the default hyperparameters of each method in MMRotate and train DOTA over 12 epochs, C19TD over 36 epochs, and HRSC2016, ICDAR2015 and MSRA-TD500 over 72 epochs.

\subsection{Main results}
Overall, our method consistently achieves SOTA performance and surpasses previous works in several cases. Qualitative detection examples of the proposed structure tensor representation (ST) on the different evaluation datasets are provided in Figure~\ref{fig:qualitative_results}. The following text provides and discusses in detail the obtained quantitative results.

\pparagraph{Results on Satellite Imagery.} We compute the standard mean average precision metrics mAP50 and mAP50:95 on all methods and report them on Table~\ref{table:dota_hrsc_results}. As shown, our approach achieves SOTA results for the DOTA dataset on OBB-supervised methods, and yields significant mAP50 improvements of +1 point on HBB-supervised methods. On HRSC2016, the structure tensor representation outperforms all other methods in both mAP50 and mAP50:95, for OBB and HBB-supervised methods.

\pparagraph{Results on Scene Text Detection and C19TD.} Table~\ref{table:text_and_covid_results} provides the mAP50 results on the oriented scene text detection data, ICDAR15 and MSRA-TD500, and the COVID19 Test dataset. As shown, the structure tensor representation outperforms other approaches on all three benchmarks within OBB-supervised approaches. In the HBB-supervised setup, our approach surpasses other encoders for ICDAR15 and MSRA-TD500 and closely follows PSC for the best score on C19TD. Notice there is a considerable drop of performance on scene text detection datasets between HBB and OBB-supervised tasks. This can be attributed to the difference in training data, where ICDAR15 and MSRA-TD500 contain considerably fewer images. When learning orientation from the redundancy in the data without explicit orientation information, it is expected that these methods are less robust in conditions of limited training data.

\subsection{Ablation study}
\paragraph{Precision of Angular Prediction.} While standard mAP provides a general indication of model performance, it does not directly reflect the precision of a network's angular predictions, as better angular prediction might not affect the true positive/false negative count. Hence, we introduce two complementary metrics that specifically evaluate orientation accuracy: the mean absolute error (MAE$_{\bm{\theta}}$) and root mean square error (RMSE$_{\bm{\theta}}$) of the predicted angle relative to the ground truth for all true positive detections based on mAP50. This way, MAE$_{\bm{\theta}}$ provides an intuitive measure of angular error in radians, while RMSE$_{\bm{\theta}}$ penalizes large errors more heavily. It should be noted that angular discontinuity and symmetry ambiguities may yield significant angle errors even when predictions are close to the ground truth. To mitigate this, we consider the $\pi$ periodicity of rectangular bounding boxes, calculating the angle error $\delta$ between prediction $\theta_{pred}$ and ground truth $\theta_{gt}$ as follows:
\begin{equation}
\label{eq:angle_error}
\delta = \min\left(
\begin{array}{l}
    \min\left(|\theta_{\text{pred}} - \theta_{\text{gt}}|, |\theta_{\text{pred}} - \theta_{\text{gt}} + \pi| \right), \\
    |\theta_{\text{pred}} - \theta_{\text{gt}} - \pi|
\end{array}
\right).
\end{equation}
MAE$_{\bm{\theta}}$ and RMSE$_{\bm{\theta}}$ are then computed using $\delta$ in their standard equations. Table~\ref{table:rmse_mae_results} compares the MAE$_{\bm{\theta}}$ and RMSE$_{\bm{\theta}}$ scores for OBB-supervised methods on the DOTA and HRSC2016 datasets. Circular objects from the DOTA dataset are excluded, i.e. \textit{baseball-diamond}, \textit{storage-tank}, \textit{roundabout}, as they may have more arbitrary orientations. Results indicate that both our approach and PSC achieve significantly lower angular errors than other methods. 

\begin{table}[t]
\centering
\begin{tabular}{ccccc}
\hline
\textbf{Dataset} & \textbf{Method} & \textbf{MAE$_{\bm{\theta}}$} & \textbf{RMSE$_{\bm{\theta}}$} \\
\hline
\multirow{8}{*}{\textbf{DOTAv1.0}} 
& Rotated RetinaNet~\cite{retinanet} & 0.075 & 0.233 \\
& Rotated FCOS~\cite{fcos} & 0.533 & 0.870 \\
& CSL~\cite{csl} & 0.066 & 0.202 \\
& GWD~\cite{wasserstein_loss} & 0.086 & 0.223 \\
& KLD~\cite{kld} & 0.547 & 0.877 \\
& KFIoU~\cite{kfiou} & 0.104 & 0.251 \\
& PSC~\cite{psc} & \textbf{\textcolor{darkgreen}{0.059}} & \textbf{\textcolor{blue}{0.196}} \\
& \textbf{ST (Ours)} & \textbf{\textcolor{blue}{0.068}} & \textbf{\textcolor{darkgreen}{0.188}} \\
\hline
\multirow{8}{*}{\textbf{HRSC2016}} 
& Rotated RetinaNet~\cite{retinanet} & 0.033 & 0.054 \\
& Rotated FCOS~\cite{fcos} & 0.023 & 0.040 \\
& CSL~\cite{csl} & 0.038 & 0.049 \\
& GWD~\cite{wasserstein_loss} & 0.025 & 0.037 \\
& KLD~\cite{kld} & 0.024 & 0.040 \\
& KFIoU~\cite{kfiou} & 0.031 & 0.048 \\
& PSC~\cite{psc} & \textbf{\textcolor{darkgreen}{0.019}} & \textbf{\textcolor{blue}{0.028}} \\
& \textbf{ST (Ours)} & \textbf{\textcolor{darkgreen}{0.019}} & \textbf{\textcolor{darkgreen}{0.027}} \\
\hline
\end{tabular}
\caption{\textbf{Angular prediction precision.} MAE$_{\bm{\theta}}$ and RMSE$_{\bm{\theta}}$ scores on DOTAv1.0 and HRSC2016 across different methods. The best score is shown in green and the second-best score in blue.\vspace{-3mm}}
\label{table:rmse_mae_results}
\end{table}

\paragraph{Effect of Anisotropy in the ST Representation.} As shown in Figure~\ref{fig:loss_plot_isotropic_issues} and discussed in Section~\ref{sec:method}, introducing anisotropy to squared objects theoretically reduces the orientation and loss issues caused by the isotropic nature of these instances. Thus, we test this in practice by comparing the results of the structure tensor representation using isotropic squared objects $(\lambda_1=\frac{w}{2}, \lambda_2=\frac{h}{2})$, and anisotropic squared objects with a 2:1 $(\lambda_1=w, \lambda_2=\frac{h}{2})$ and 4:1 $(\lambda_1=2w, \lambda_2=\frac{h}{2})$ aspect ratio, respectively. We report the results on DOTA and HRSC  in Table~\ref{table:ablation_anisotropy}, in which we can observe how anisotropic structure tensors considerably improve angle error and mAP50. While a 4:1 anisotropic structure tensor shows the lowest MAE$_{\bm{\theta}}$, the 2:1 tensor achieves the best mAP50 with a minimal MAE$_{\bm{\theta}}$ increase that corresponds to 0.34 degrees. This behavior aligns with the idea that, the more anisotropic the representation, the easier it is for the network to estimate the angle as there is a clear principal direction and a larger penalization of wrong predictions in the loss. This seems to occur until a point is reached, where the extreme anisotropy in the tensor deviates from the real object shape, and mAP scores are affected as a result. Therefore, we keep the 2:1 anisotropic structure tensor in all other experiments.

\begin{table}[t]
\centering
\resizebox{\columnwidth}{!}{
\begin{tabular}{cccccc}
\hline
\textbf{Dataset} & \textbf{Anisotropy} & \textbf{MAE$_{\bm{\theta}}$} & \textbf{RMSE$_{\bm{\theta}}$} & \textbf{mAP50} \\
\hline
\multirow{3}{*}{\textbf{DOTAv1.0}} 
& Isotropic & 0.083 & 0.200 & 70.4 \\
& 2:1 & 0.068 & 0.188 & \textbf{71.3} \\
& 4:1 & \textbf{0.062} & \textbf{0.185} & 69.6 \\
\hline
\multirow{3}{*}{\textbf{HRSC2016}} 
& Isotropic & 0.021 & 0.032 & 89.9 \\
& 2:1 & 0.019 & \textbf{0.027} & \textbf{90.3} \\
& 4:1 & \textbf{0.018} & \textbf{0.027} & 90.1 \\
\hline
\end{tabular}
}
\caption{\textbf{Comparison of structure tensor representations with different levels of anisotropy}. Angle MSE ($MAE_{\theta}$), angle RMSE ($RMSE_{\theta}$), and mAP50 are shown for DOTAv1.0 and HRSC2016. The anisotropic structure tensor with a 2:1 aspect ratio yields the best mAP50 scores, with minimal angle error.}
\label{table:ablation_anisotropy}
\end{table}
\paragraph{Computational Complexity.} Lastly, we analyze the computational complexity of our approach and other methods. To this aim, we report in Table~\ref{table:computational_complexity} the FLOPs (in GFLOPs) and the number of parameters of all evaluated methods. As shown, the structure tensor representation stays on the lower end on both metrics, providing a convenient balance between performance and computational complexity. 
\begin{table}[t]
\centering
\resizebox{\columnwidth}{!}{
\begin{tabular}{lccc}
\hline
\textbf{Method}   & \textbf{Architecture} & \textbf{FLOPs} & \textbf{Params (M)} \\ \hline
rotated FCOS      & FCOS              & 206.92                  & 31.92               \\
CSL               & FCOS              & 215.91                  & 32.34               \\
KLD               & FCOS              & 206.01                  & 31.92               \\
PSC               & FCOS              & 207.16                  & 31.93               \\
rotated RetinaNet & RetinaNet         & 209.58                  & 36.13               \\
GWD               & RetinaNet         & 209.58                  & 36.13               \\
KFIoU             & RetinaNet         & 215.92                  & 36.42               \\ 
\textbf{ST (Ours)}  & FCOS                 & 207.01                  & 31.93         \\ \hline
\end{tabular}
}
\caption{\textbf{Comparison of models in terms of parameters (in millions) and computational complexity (FLOPs)}. All FLOPs are measured in GFLOPs. ResNet-50 was used as the backbone for all measurements. Our method remains on the lower end in terms of both FLOPs and parameter count compared to other models.\vspace{-6mm}}

\label{table:computational_complexity}
\end{table}

\section{Conclusion}
\label{sec:conclusion}
In this article, we present a novel angle representation for oriented object detection that effectively addresses angular discontinuities and symmetric ambiguities by encoding orientation as a structure tensor. Our method is straightforward, efficient, and does not require hyperparameter tuning, unlike many existing approaches. We evaluate our representation across several datasets, including satellite imagery, scene text detection, and symmetric objects, comparing it against state-of-the-art methods. Our approach yields significant mAP results, consistently achieving SOTA performance across different benchmarks. In some cases, we even outperform previous methods in both OBB and HBB-supervised tasks. Additionally, we conduct ablation studies to assess the impact of anisotropy in the structure tensor representation, and demonstrate that our approach provides precise angular prediction at a low computational cost.


{
    \small
    \bibliographystyle{ieee_fullname}
    \bibliography{egbib}
}


\end{document}